%% file: main.tex
\documentclass[letterpaper, 10 pt, conference]{template/ieeeconf}
\IEEEoverridecommandlockouts
\usepackage{bm}
\usepackage{algorithm}
\usepackage{algpseudocode}
\usepackage{amsmath}
\usepackage{amsfonts}
\usepackage{breqn}
\usepackage{graphicx} 
\usepackage{subfig} 
\usepackage{multirow}
\usepackage{nicefrac}
\usepackage{dblfloatfix}  
\usepackage[export]{adjustbox}  
\usepackage{hyperref}
\usepackage[dvipsnames]{xcolor}
\usepackage{todonotes}
\usepackage[utf8]{inputenc}
\usepackage{booktabs}

\newcommand{\x}{\textbf{x}}

\newcommand{\bS}{\bm{\Sigma}}

\newcommand{\etal}{\textit{et al. }}

\title{\LARGE \bf Sparse-to-Continuous: Enhancing Monocular Depth \\ Estimation using Occupancy Maps}

\author{Nícolas dos Santos Rosa$^{1}$, Vitor Guizilini$^{2}$, and Valdir Grassi Jr$^{1}$ %
\thanks{The authors are with the São Carlos School of Engineering$^{1}$ at University of São Paulo (USP, Brazil) and the Toyota Research Institute$^{2}$, Los Altos, CA. Emails: {\tt\small \{nicolas.rosa; vgrassi\}@usp.br, vitor.guizilini@tri.global}\newline
Code available at: \href{https://github.com/nicolasrosa/Sparse-to-Continuous}{\textcolor{purple}{https://github.com/nicolasrosa/Sparse-to-Continuous}}}%
}

\begin{document}
\bstctlcite{IEEEexample:BSTcontrol}

\maketitle
\thispagestyle{empty}
\pagestyle{empty}

\input{00abstract}
\input{01introduction}
\input{02related_work}
\input{03methodology}
\input{04experiments}
\input{05conclusion}

\input{06acknowledgment}
\input{07references}

\end{document}

%% file: 00abstract.tex
\begin{abstract}

This paper addresses the problem of single image depth estimation (SIDE), focusing on improving the quality of deep neural network predictions. In a supervised learning scenario, the quality of predictions is intrinsically related to the training labels, which guide the optimization process. For indoor scenes, structured-light-based depth sensors (e.g.\ Kinect) are able to provide dense, albeit short-range, depth maps. On the other hand, for outdoor scenes, LiDARs are considered the standard sensor, which comparatively provides much sparser measurements, especially in areas further away. Rather than modifying the neural network architecture to deal with sparse depth maps, this article introduces a novel densification method for depth maps, using the Hilbert Maps framework. A continuous occupancy map is produced based on 3D points from LiDAR scans, and the resulting reconstructed surface is projected into a 2D depth map with arbitrary resolution. Experiments conducted with various subsets of the KITTI dataset show a significant improvement produced by the proposed Sparse-to-Continuous technique, without the introduction of extra information into the training stage.

\end{abstract}

%% file: 01introduction.tex
\section{Introduction}
\label{sec:introduction}

Robotic platforms have been increasingly present in our society, performing progressively more complex activities in the most diverse environments. One of the driving factors behind this breakthrough is the development of sophisticated perceptual systems, which allow these platforms to understand the environment around them as well as -- or better than -- humans. For this, these platforms should be able to extract depth information from the environments where they are inserted. This work introduces a preprocessing technique that benefits the training of deep convolutional networks used to retrieve depth information from monocular images.

Nowadays, sensors allow the large-scale capture of three-dimensional information, and amongst them, the most commonly used are rangefinders using LiDAR technology~\cite{schwarz2010lidar}. Nonetheless, these sensors can be extremely expensive depending on the range and level of detail required by the application. Since it is also possible to reconstruct 3D structures from 2D observations of the scene~\cite{Rezende2016}, visual systems have been employed as an alternative. This is motivated due to their reduced cost and size, while also being able to perceive colors. However, estimating depths from 2D images is a challenging task and it is described as an ill-posed problem. The reason for this is that the observed images may be resultant of several possible projections from the actual real-world scene~\cite{Eigen2014}. This problem has been extensively studied in Stereo Vision~\cite{Chen2015stereo, Zbontar2015, Mayer2016, Luo2016} and Single Image Depth Estimation (SIDE)~\cite{Eigen2014, Liu2015, Godard2016, Fu2018}. In this work, we focus on the second approach, since it only requires one camera.

\begin{figure}[!t]
	\centering
	\includegraphics[width=0.45\textwidth]{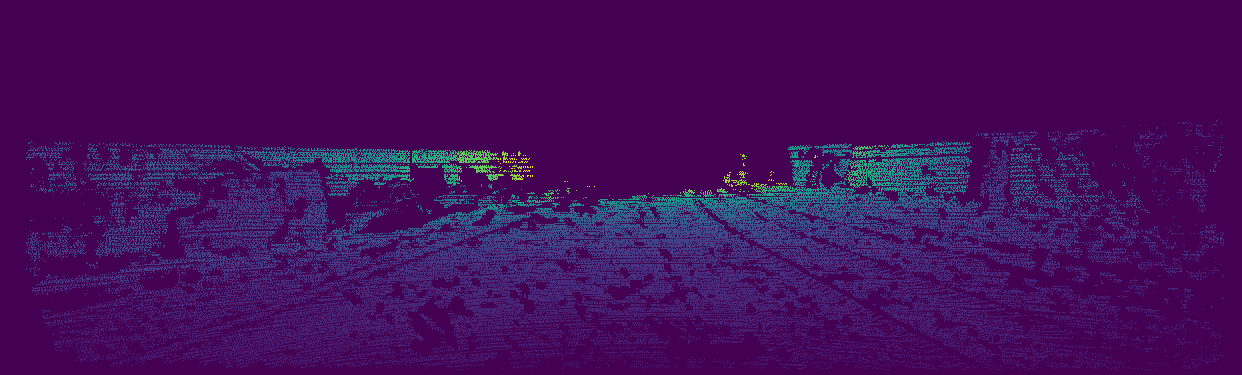}\\
	\vspace{2pt}
	\includegraphics[width=0.45\textwidth]{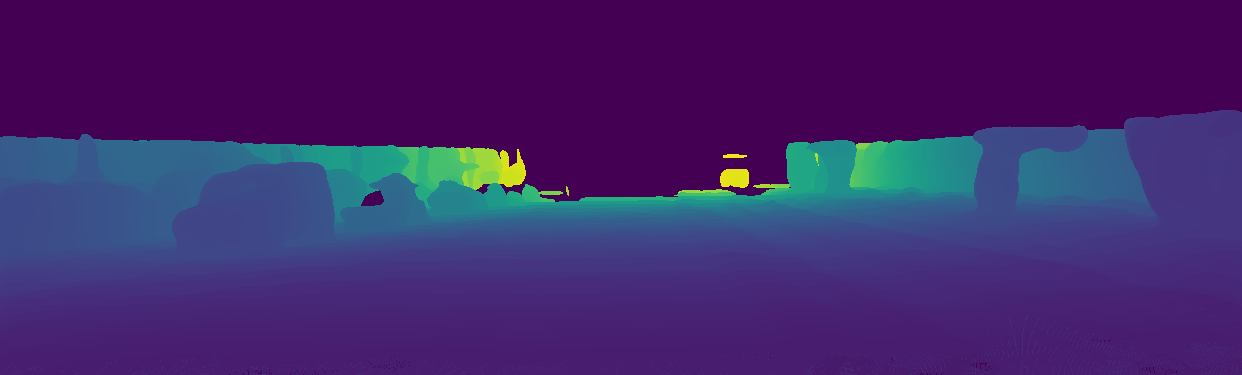}
	\caption{Sparsity comparison between the (a) \textit{KITTI Depth} (semi-dense) and (b) \textit{KITTI Continuous} (dense, ours) datasets, respectively. Warmer and colder colors represent larger and smaller distances, respectively.}
	\vspace{-0.3cm}
	\label{fig:kitti_datasets}
\end{figure}

Deep Convolutional Networks (CNNs) have had a deep impact on how recent works address the SIDE task, with significant improvements in the accuracy and level of details present in depth maps. Many of these methods model the monocular depth estimation task as a regression problem and are supervised. They often use sparse depth maps as ground-truth, since these are readily available from other sensors (i.e.\ LiDAR rangefinders).

However, the degree of sparsity present in these maps is very high. For instance, in a $375 \times 1242$ image from the KITTI Depth dataset, 84.29\% of its pixels do not contain valid information. For these pixels, the LiDAR fails to acquire data due to light beams that didn't return to the sensor (e.g.\ sky). Probably because they couldn't be projected to the camera's image sensor, or maybe the sensor is just sparse and the measured values don't fall in those pixels. For the effective training of deep neural networks, one of the palliatives found to overcome this lack of information is to use datasets that provide a large number of examples. Strictly for outdoor depth estimation tasks, the KITTI Raw Data~\cite{Geiger2013IJRR} and KITTI Depth~\cite{Uhrig2017} fit the above-mentioned requirement. However, they also have the problem of high sparsity on depth maps. Some recent works also propose the use of secondary information (i.e.\ low-resolution depth maps, normal surfaces, semantic maps), associating them to the RGB images as extra inputs~\cite{Ma2018, Liao2017, zhang2018deep}. Other works focused on the development of network architectures that are more suitable for processing sparse information~\cite{Uhrig2017, hua2018normalized}.

Similar existing works have also proposed the use of rendered depth images from synthetic datasets, which are continuous~\cite{Mancini2017towards, Mancini2018jmod}. However, the generalization power of networks trained using this type of dataset is still questioned~\cite{Uhrig2017}. This is mainly due to the existing gap in terms of the degree of realism between virtual environments and real-world scenes.

In this work, we address the SIDE task and propose the use of a densification method to interpolate regions with valid information (i.e.\ from raw sparse LiDAR measurements). Next, we generate continuous depth images, which then serve to train a deep convolutional network in a supervised manner. In our case, occupation models are used as the densification technique and a ResNet-based architecture is adopted to map RGB images to continuous depth maps. The same concept can also be applied to other in-painting techniques and network architectures. The main contribution of this paper is to show that, by exploiting these continuous images, we can make the network to improve the quality of predictions at test time. Finally, to demonstrate the benefits of training deep convolutional networks using our proposed method, we compare the obtained estimates when training in two different datasets with varying levels of sparsity, as illustrated in Figure~\ref{fig:kitti_datasets}.

To produce the occupancy models necessary for continuous projections, we employ the Hilbert Maps framework~\cite{RamOtt2016}, due to its efficient training and query properties, and scalability to large datasets. Previously, Hilbert maps were strictly related to \textit{Motion Planning} problems as a way to improve object recognition and reinforcement learning tasks. Instead, this work presents a novel application for this technique, in which it is used as a densification method to benefit directly the SIDE task.


%% file: 02related_work.tex
\section{Related Work}
\label{sec:related_work}

\textbf{Depth Estimation} from a single image is an ill-posed problem, since the observed image may be generated from several possible projections from the actual real-world scene~\cite{Eigen2014}. Besides, it is inherently ambiguous, as the proposed methods attempt to retrieve depth information directly from color intensities~\cite{Laina2016}. Besides all the presented adversities, other tasks such as obstacle detection, semantic segmentation, and scene structure highly benefit from the presence of depth estimates~\cite{Cao2018}, which makes this task particularly useful.

Previous approaches relied on handcrafted features, manually selected to have useful properties, and had strong geometrical assumptions~\cite{Hoiem2005automatic, Hedau2010}. More recently, probabilistic graphical models and deep convolutional networks have been employed for retrieving contextual information and extracting visual cues and multi-scale hierarchical features present in the scene. Commonly, the monocular depth estimation problem is modeled as a regression problem, whose parameters are optimized based on the minimization of a cost function,  often using sparse depth maps as ground-truth for supervised learning.

Early works employed techniques such as Markov Random Fields (MRFs)~\cite{Saxena2006, Saxena2009} and Conditional Random Fields (CRFs)~\cite{Liu2014} to perform this task. More recently, deep learning concepts have also been used to address the SIDE problem, where deep convolutional neural networks (CNNs) are responsible for extracting the visual features~\cite{Eigen2014, Eigen2015, Liu2015, Laina2016}. Hybrid architectures, which jointed both techniques, were also employed~\cite{Liu2015, Wang2015}. The success of these techniques highly impacted how subsequent works began to address the SIDE task, which in turn significantly improved the accuracy of estimates and the level of details present in depth maps~\cite{Ma2018, Fu2018}.

In parallel to the supervised learning approach, some works focus on minimizing photometric reconstruction errors between the stereo images~\cite{Garg2016, Godard2016} or video sequences~\cite{Zhou2017}, which allow them to be trained in an unsupervised way (i.e.\ without depth estimates as ground-truth).

\textbf{Depth Map Completion} has been widely studied in computer vision and image processing applications and deals with decreasing the sparsity level of depth maps. Monocular Depth Estimation differs from it as it seeks to directly approximate RGB images to depth maps. In summary, existing Depth Map Completion methods seek to predict distances for pixels where the depth sensor doesn't have information. Currently, there are two types of approaches associated with this problem.


The first one, \textit{non-Guided Depth Upsampling}, aims to generate denser maps using only sparse maps obtained directly from 3D data or SLAM features. These methods resemble those proposed in the Depth Super-Resolution task~\cite{Uhrig2017}, where the goal is to retrieve accurate high-resolution depth maps. More recently, deep convolutional neural networks have also been employed in super-resolution for both image~\cite{Yang2010, Dong2014} and depth~\cite{Riegler2016, song2016deep} applications. Other works focus on inpainting the missing depth information, e. g., Uhrig \etal~\cite{Uhrig2017} employed sparse convolutional layers to process irregularly distributed 3D laser data. As pointed by Zhang and Funkhouser~\cite{zhang2018deep}, methods predicting depth when trained only on raw information usually do not perform too well.

The second approach, \textit{Image Guided Depth Completion}, suggests incorporating some kind of guidance for achieving superior performance, e. g. to use sparse maps and RGB images of the scene (RGB-D data) as inputs. Besides low-resolution sparse samples obtained from low-cost LiDAR or SLAM features~\cite{Ma2018, Weerasekera2018}, other auxiliary information can also be employed, such as semantic labels~\cite{Schneider2016depth}, 2D laser points~\cite{Liao2017}, normal surface, and occlusion boundary maps~\cite{zhang2018deep}.

%% file: 03methodology.tex
\section{Methodology}
\label{sec:methodology}

\subsection{Occupancy Maps}
\label{sec:occupancy_maps}

A common way to store range-based sensor data is through the use of pointclouds, which can be projected back into a 2D plane to produce depth images, containing distance estimates for all pixels that have a corresponding world point. Assuming a rectified camera projection matrix $\mathbf{P}_{rect} \in \mathbb{R}^{3 \times 4}$, a rectifying rotation matrix $\mathbf{R}_{rect} \in \mathbb{R}^{3 \times 3}$ and a rigid body transformation matrix from camera to range-based sensor $\mathbf{T}_{range}^{cam} \in \mathbb{R}^{4 \times 4}$, a 3D point $\textbf{p}$ can be projected into pixel $\textbf{u}$ as such:
\begin{equation}
\textbf{u} = \textbf{P}_{rect} \textbf{R}_{rect} \textbf{T}_{range}^{cam} \textbf{p} \hspace{0.05cm} .
\end{equation}
An example of this projection can be seen in Figure~\ref{fig:kitti_datasets}a, where we can see the sparsity generated by directly projecting pointcloud information, most notably in areas further away from the sensor. Spatial dependency modeling is a crucial aspect in computer vision, and the introduction of such irregular gaps can severely impact performance. Because of that, here we propose projecting not the pointcloud itself, but rather its occupancy model, as generated by the Hilbert Maps (HM) framework~\cite{RamOtt2016}. This methodology has recently been successfully applied to the modeling of large-scale 3D environments~\cite{GuiRam2016}, producing a continuous occupancy function that can be queried at arbitrary resolutions. Assuming a dataset $\mathcal{D}=\{ \textbf{x}_i , y_i \}_{i=1}^N$, where $\textbf{x}_i \in \mathcal{R}^3$ is a point in the three-dimensional space and $y_i = \{ -1 , +1 \}$ is a classification variable that indicates the occupancy property of $\textbf{x}_i$, the probability of non-occupancy for a query point $\textbf{x}_*$ is given by:
\begin{equation}
p(y_* = -1|\Phi(\textbf{x}_*),\textbf{w}) = \frac{1}{ 1 + \exp \left( \textbf{w}^T\Phi(\textbf{x}_*) \right) },
\end{equation}
where $\Phi(\textbf{x}_*)$ is the feature vector and $\textbf{w}$ are the weight parameters, that describe the discriminative model $p(y|\textbf{x},\textbf{w})$. We employ the same feature vector from~\cite{GuiRam2016}, defined by a series of squared exponential kernel evaluations against an inducing point set $\mathcal{M} = \{ \mathcal{M}_i\}_{i=1}^M = \{ \bm{\mu}_i,\bS_i \}_{i=1}^M $, obtained by clustering the pointcloud and calculating mean $\bm{\mu}$ and variance $\bm{\Sigma}$ estimates for each subset of points:
\begin{align} \label{eq:lard} \small
\Phi(\textbf{x},\mathcal{M}) &= 
\Big[ 
k(\textbf{x},\mathcal{M}_1) , 
\dots , 
k(\textbf{x},\mathcal{M}_M)
\Big] \\
k(\x,\mathcal{M}_i) &= \exp \left( - \frac{1}{2} ( \x - \bm{\mu}_i )^T \bS_i^{-1} ( \x - \bm{\mu}_i ) \right).
\end{align}
Clustering is performed using the Quick-Means algorithm proposed in~\cite{GuiRam2017rss}, due to its computational efficiency and ability to produce consistent cluster densities. However, this algorithm is modified to account for variable cluster densities within a function, in this case the distance $d$ from origin. This is achieved by setting $r_i = r_o = \tau \cdot f(d)$, where $r_i$ and $r_o$ are the inner and outer radii used to define cluster size and $\tau$ is a scaling constant. The intuition is that areas further from the center will have fewer points, and therefore larger clusters are necessary to properly interpolate over such sparse structures. The trade-off for this increase in interpolative power is loss in structure details, since a larger volume will be modeled by the same cluster. The optimal weight parameters $\textbf{w}$ are calculated by minimizing the following negative-likelihood loss function:
\begin{equation} \small
\mathcal{L}(\textbf{w}) =\sum_{i=1}^N \log \left( 1 + \exp\left( -y_i\textbf{w}^T \Phi(\textbf{x}_i) \right) \right) + R(\textbf{w}) \label{eq:nnl}.
\end{equation}
where $R(\textbf{w})$ is a regularization function such as the elastic net~\cite{Zou2005regularization}. Once the occupancy model has been trained, it can be used to produce a reconstruction of the environment, and each pixel is then checked for collision in the 3D space, producing depth estimates. An example of reconstructed depth image is depicted in Figure~\ref{fig:kitti_datasets}c, where we can see that virtually all previously empty areas were filled by the occupancy model, while maintaining spatial dependencies intact (up to the reconstructive capabilities of the HM framework).

\subsection{Continuous Depth Images}
\label{sec:continuous_depth_images}

When datasets do not provide ground truths directly, it is still possible to obtain them using 3D LiDAR scans and extrinsic/intrinsic parameters from the RGB cameras. In this case, a sparse depth image can be generated by directly projecting the cloud of points of the scene to the image plane of the visual sensor~\cite{Eigen2014, Godard2016}. Continuous depth images, in turn, can be obtained by interpolating the measured points into continuous surfaces prior to the projection. In this work, the Hilbert Maps framework is used on the LiDAR scans to generate these surfaces. After restricting the continuous map to the region under the camera's field of view, we project the remaining depth values in the image plane.

\subsection{Data Augmentation}
\label{subsec:data_augmentation}

Two types of random online transformations were performed, thus artificially increasing the number of training data samples.

\textbf{Flips}: The input image and the corresponding depth map were flipped horizontally with 50\% probability. 

\textbf{Color Distortion}: Adjusts the intensity of color components on an RGB image randomly. The order of the following transformations was also chosen randomly: \textit{Brightness}, \textit{Saturation}, \textit{Hue} and \textit{Contrast}.

As pointed out by~\cite{Eigen2014}, the world-space geometry of the scene is not preserved by image scaling and translation. Therefore, we opted for not using these transformations. We believe that aggressive color distortions prevent the network from becoming biased in relating pixel intensity to depth values, thus focusing on learning the scene's geometric relationships.

\subsection{Loss Functions}
\label{sec:loss_functions}

We employed three different loss functions for adjusting the internal parameters of the presented deep neural network: MSE~\cite{Laina2016}, Eigen~\cite{Eigen2015}, and BerHu~\cite{Laina2016, owen2007robust, Zwald2012}. The motivation behind this is simply to determine which one is more suitable for approximating the outputs ($y$) to the reference values ($y^*$) for the $i$-th pixel. The mathematical expressions for each one are presented as follows:

\subsubsection{Squared Euclidian Norm (mse)}



\begin{equation}
    \label{eq:squared_euclidian_loss}
    \mathcal{L}_2(y - y^*) = \left \|  y - y^* \right \|_2^2 = \sqrt{\sum_{i}^{n} (y_i - y_i^*)^2}
\end{equation}

\subsubsection{Scale Invariant Mean Squared Error (eigen)}



\begin{dmath}
    \label{eq:scale_inv_loss_with_grads}
    \mathcal{L}_{eigen_{grads}}(y, y^*) = \frac{1}{n} \sum_{i} d_i^2 - \frac{\lambda}{n^2} \left(\sum_i d_i \right)^2 + \frac{1}{n} \sum_{i} [(\nabla_x d_i)^2+(\nabla_y d_i)^2],
\end{dmath}

\vspace{-0.25cm}

$$d_i = \log \ y_i - \log \ y_i^*, \quad\lambda = \frac{1}{2}$$

\subsubsection{Adaptive BerHu Penalty (berhu)} 


\begin{equation}
    \label{eq:berhu_loss}
    \mathcal{L}_{berhu}(x) = \left\{\begin{matrix} \left | x \right |\quad\ \quad\left | x \right |  \leq c,\\ 
    \frac{x^2+c^2}{2c}\quad \left | x \right | > c
    \end{matrix}\right. ,\quad x = y - y^*
\end{equation}

\vspace{-0.10cm}

$$c = \frac{1}{5}\ \text{max}_i(\left | y_i-y_i^* \right |)$$

In the \textit{BerHu} loss (Equation~\ref{eq:berhu_loss}), the penalization adapts according to how far the predictions are from the reference depths, where small values are subject to the $\mathcal{L}_1$ norm, whereas high values, to the $\mathcal{L}_2$ norm.

\vspace{-0.25cm}
\subsection{Network Architecture}
\label{subsec:network_architecture}

\begin{figure*}[htp]
\centering
\includegraphics[width=0.8\textwidth]{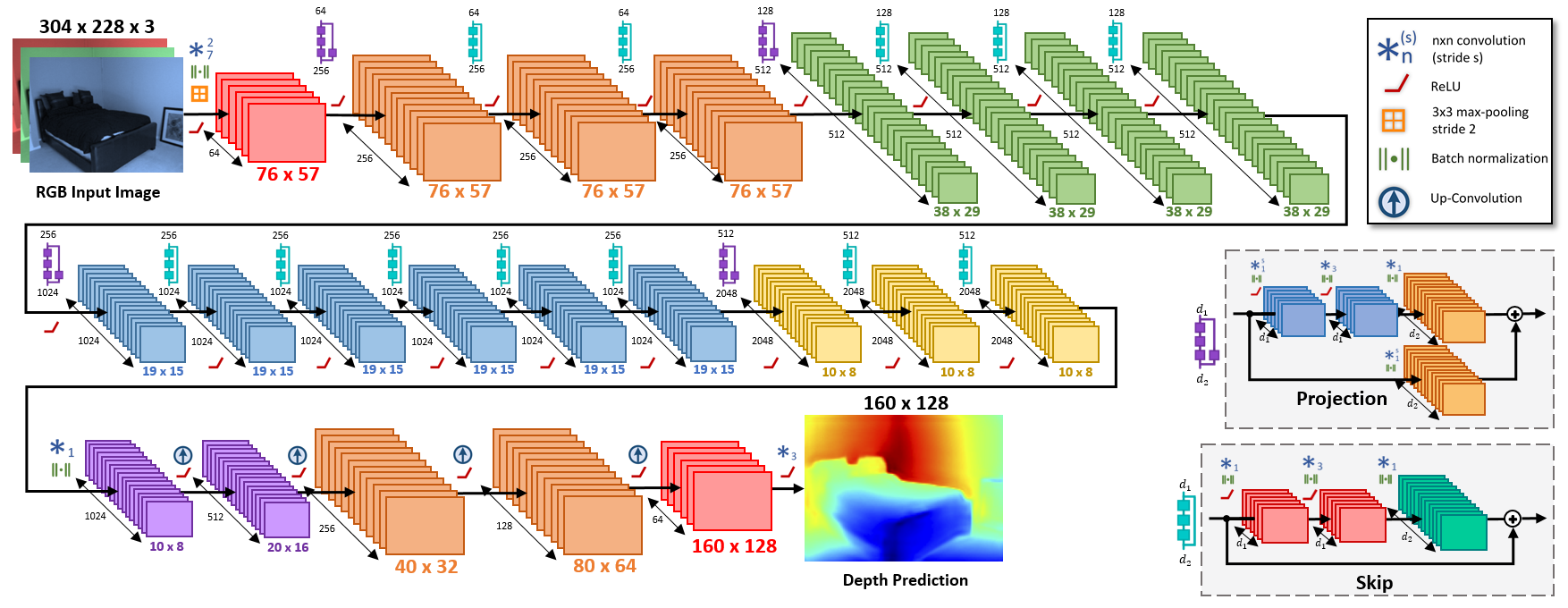}
\caption{Network architecture. The used architecture was proposed by Laina \etal~\cite{Laina2016}, which is inspired on the ResNet-50, but its fully-connected layers were replaced by upsampling blocks. Differently from the original authors, we changed the training framework to make the proposed network predicts meters instead of distances in log-space. Figure adapted from Laina \etal~\cite{Laina2016}.}
\label{fig:resnet_laina}
\end{figure*}

In this work, we used a \textit{Fully Convolutional Residual Network} (FCRN) as proposed by Laina \etal~\cite{Laina2016}. This network was selected because it presents a smaller number of trainable parameters, thus requiring a smaller number of images to be trained without losing performance. Also, the residual blocks present in the architecture allow the construction of a deeper model capable of predicting more accurate and higher-resolution output maps. More specifically, the FCRN (Figure~\ref{fig:resnet_laina}) is based on the ResNet-50 topology, but the fully-connected layers have been replaced by a set of residual upsampling blocks, also referred to as \textit{up-projections}, which are layers responsible for deconvolving and retrieving spatial resolution of feature maps. This network was trained end-to-end in a supervised way, but unlike the original authors, we modified its output to predict distances in meters rather than distances in log-space. The network uses RGB images of $304 \times 228$ pixels as inputs for adjusting the 63M trainable parameters and provides an output map with $128 \times 160$ pixels.

%% file: 04experiments.tex
\section{Experimental Results}
\label{sec:experiments}

\subsection{Experiment Description}
\label{subsec:experiment_description}

To demonstrate the effectiveness of the proposed method, an experimentation baseline was developed, in which the same network architecture and the same training hyperparameters were used. That created a testing environment where only differences of sparsity are evaluated. Thus, the results obtained demonstrate how the different levels of sparsity (semi-dense and dense) in the datasets considered here influence the quality of final predictions.

\subsection{Implementation details}
\label{subsec:implementation_details}

We implemented the network using Tensorflow~\cite{Abadi2016} and the Adam algorithm as optimizer~\cite{Kingma2014adam}. Our models were trained using an NVIDIA Titan X GPU with 12 GB memory. We used a batch size of 4 and 300000 training steps. The initial learning rate value was 0.0001, reducing 5\% every 1000 steps. Besides learning decay, we also employed a dropout of 50\% and $\mathcal{L}_2$ normalization as regularization~\cite{srivastava2014, Goodfellow2015}.

\subsection{Datasets}
\label{subsec:datasets}

Two different datasets are considered in this work: \textit{KITTI Depth} (semi-dense), and \textit{KITTI Continuous} (dense), including frames from the ``city'', ``residential'', ``road'', ``campus'' and ``person'' sequences. The dense one was generated from \textit{KITTI Raw}. Typically, the resolution of the used RGB and depth maps images is $375 \times 1242$ pixels.\\

\noindent$\bullet$ \textit{KITTI Depth:} This dataset is paired with scenes presented in the \textit{KITTI Raw} dataset~\cite{Geiger2013IJRR} and consists of 92750 semi-dense depth maps (ground truth), subdivided into 85898 and 6852, for the train and validation sets, respectively. The depth images were obtained by accumulating 11 lasers scans, whose outliers were removed by enforcing consistency between the LiDAR points and the reconstructed depth maps, generated by semi-global matching (SGM)~\cite{Uhrig2017}. In this work, we used both left and right images.\\

\noindent$\bullet$ \textit{KITTI Continuous:} Unlike the procedure performed in \textit{KITTI Depth} to make the depth images less sparse, which consisted of accumulating different scans of the laser sensor, we used an occupancy model to make the depth maps denser. In other words, the goal was to increase the number of valid pixels available in depth images for training. In this sense, this alternative requires a smaller number of training images than other techniques that use datasets with semi-dense ground truth information. The \textit{KITTI Continuous} is also based on 3D Velodyne point clouds, but first we interpolate its measurements as surfaces to generate the continuous depth images (more details in section~\ref{sec:continuous_depth_images}).\\

Due to limitations in our preprocessing framework, we randomly generated a smaller number of images \textit{KITTI Continuous} datasets. More specifically, 27,817 and 2,911 samples (RGB Image + Depth Image) were created for the train and validation sets, respectively. Even with fewer images, we use the \textit{KITTI Depth} official train/validation split as a lookup table, strictly maintaining the original pairs' separation, i.e.\  pairs that used to be part of the train/validation subset continued to belong to it.

The resulting number of pairs for each dataset is presented in Table~\ref{tab:number_of_images}. On average, the number of valid pixels available on the \textit{KITTI Depth} datasets at 128x160 resolution represents 71.84\% of the number of points available on the \textit{KITTI Continuous} dataset, i.e.\  1.39 times smaller.




\begin{table}[h]
\centering
\caption{Number of image pairs used in train/validation sets for each dataset.}
\label{tab:number_of_images}
\resizebox{0.5\textwidth}{!}{%
\begin{tabular}{lccccl}
\hline
\multicolumn{1}{c}{\multirow{2}{*}{Dataset}} & \multirow{2}{*}{\textbf{Train}} & \multirow{2}{*}{\textbf{Valid}} & \multirow{2}{*}{\textbf{Total}} & \multicolumn{2}{c}{\begin{tabular}[c]{@{}c@{}}\textbf{Average percentage}\\\textbf{of valid pixels}\end{tabular}} \\
\multicolumn{1}{c}{} &  &  &  & 375x1242 & \multicolumn{1}{c}{128x160} \\ \hline
KITTI Depth & 85898 & 6852 & 92750 & 15.71\% & 45.98\% \\
KITTI Continuous & 27817 & 2911 & 30728 & 62.39\% & 64.00\% \\ \hline
\end{tabular}
} 
\end{table}

\subsection{Benchmark Evaluation}
\label{sec:benchmark_evaluation}

In this section, we perform the benchmark evaluation of our method trained on the presented datasets and compare them with existing works. The \textit{KITTI Raw} dataset does not have an official train/test split, so Eigen \etal subdivided the available images into 33,131 for training and 697 for testing~\cite{Eigen2014}.
Since this test subset is too sparse for accurate measurement of predictive quality in depth estimates, we used the \textit{KITTI Depth} as an alternative (652 semi-dense depth images) to evaluate depth estimation methods. 
 The network predictions were resized to the original depth map size using bilinear upsampling, and then compared to its corresponding ground-truth depth maps according to the \textit{Eigen Split based on KITTI Depth}.\\

\noindent\textbf{Eigen Split based on KITTI Depth (semi-dense)}: For a fair evaluation of the methods considered here, we chose a test split that provided a higher number of evaluation points, since our technique improves the prediction quality not only for sparse points of the original depth maps -- which are accounted for in metrics -- but also for the scene as a whole. This modification makes it possible to further highlight the benefits of point cloud densification when analyzing the impacts of using the sparse and continuous depth maps. In summary, this test split is aligned with the \textit{Eigen Split}, but uses the corresponding depth images from \textit{KITTI Depth} dataset. 
All the results presented in this paper were obtained through the above-mentioned test set.

\subsection{Evaluation Metrics}
\label{subsec:evaluation_metrics}

Since the final results are generally a set of predictions of the test set images, qualitative (visual) analysis may be biased and not sufficient to say if one approach is better than another. This way, several works use the following metrics to evaluate their methods and thus compare them with other techniques in the literature~\cite{Eigen2014, Li2015, Cao2018}:

\vspace{2mm}

\textbf{Threshold ($\delta$):} \% of $y_i$ s.t. $\max\left(\frac{y_i}{y_i^*},\ \frac{y_i^*}{y_i}\right) = \delta < thr$

\textbf{Abs Relative Difference:} $ \frac{1}{\left | T \right |}\sum_{y\ \epsilon\ T} \frac{ \left | y_i-y_i^* \right|}{y_i^*} $

\textbf{Squared Relative Difference: :} $\frac{1}{\left | T \right |}\sum_{y\ \epsilon\ T} \frac{\left \| y_i-y_i^* \right \|^2}{y_i^*}$


\textbf{RMSE (linear):} $\sqrt{\frac{1}{\left | T \right |}\sum_{y\ \epsilon\ T} \left \| y_i-y_i^* \right \|^2}$

\textbf{RMSE (log):} $\sqrt{\frac{1}{\left | T \right |}\sum_{y\ \epsilon\ T} \left \| \log\ y_i-\log\ y_i^* \right \|^2}$
\vspace{2mm}

\noindent where $T$ is the number of valid pixels in all evaluated images. In addition, to compare our results with other works, we also use the evaluation protocol of restricting ground-truth depth values and predictions to a range, in this case, the $0 - 50\ m$ and $0 - 80\ m$ intervals. 

\subsection{Results}
\label{subsec:results}

\begin{table*}[thbp]
\centering
\caption{\textbf{Performance on KITTI}. Comparison between state-of-the-art methods and the best results obtained from our method when trained on the proposed datasets, with different sparsity levels in the ground-truth depth images, loss functions and evaluated on the \textit{Eigen Split based on KITTI Depth}~\cite{Eigen2014}. The state-of-art-results are the ones presented by Lee \etal~\cite{lee2019big}. All methods were evaluated on the central crop proposed by Garg \etal~\cite{Garg2016}.}

\label{tab:sparsity_analysis_eigen_kitti_depth_split_results}
\resizebox{\textwidth}{!}{%
\begin{tabular}{clccccccc}
\toprule
& & \textbf{Abs Rel} & \textbf{Sqr Rel} & \textbf{RMSE} & \textbf{RMSE (log)} & $\bm{\delta < 1.25}$ & $\bm{\delta < 1.25^2}$ & $\bm{\delta < 1.25^3}$ \\
\textbf{Methods} & \textbf{cap} & \multicolumn{4}{c}{lower is better} & \multicolumn{3}{c}{higher is better}\\
\midrule

Ours (KITTI Depth, semi-dense, \textit{berhu}) & $0 - 50\ m$ & 0.617 & 5.484 & 10.118 & 1.189 & 0.004 & 0.077 & 0.250\\
Ours (KITTI Continuous, dense, \textit{berhu}) & $0 - 50\ m$ & 0.119 & 0.444 & 3.097 & 0.180 & 0.893 & 0.974 & 0.990\\
Godard \etal (MonoDepth)~\cite{Godard2016} & $0 - 50\ m$ & 0.076 & 0.334 & 2.613 & 0.121 & 0.929 & 0.985 & 0.996 \\
Fu \etal (DORN)~\cite{Fu2018} & $0 - 50\ m$ & 0.078 & 0.273 & 2.184 & 0.115 & 0.944 & 0.987 & 0.995 \\
Lee \etal (BTS)~\cite{lee2019big} & $0 - 50\ m$ & \textbf{0.060} & \textbf{0.182} & \textbf{2.005} & \textbf{0.092} & \textbf{0.959} & \textbf{0.994} & \textbf{0.999} \\

\midrule

Ours (KITTI Depth, semi-dense, \textit{berhu}) & $0 - 80\ m$ & 0.619 & 6.252 & 12.394 & 1.198 & 0.004 & 0.075 & 0.246\\
Ours (KITTI Continuous, dense, \textit{berhu}) & $0 - 80\ m$ & 0.123 & 0.641 & 4.524 & 0.199 & 0.881 & 0.966 & 0.986\\
Godard \etal (MonoDepth)~\cite{Godard2016} & $0 - 80\ m$ & 0.081 & 0.487 & 3.687 & 0.131 & 0.919 & 0.982 & 0.995 \\
Fu \etal (DORN)~\cite{Fu2018} & $0 - 80\ m$ & 0.081 & 0.337 & 2.930 & 0.121 & 0.936 & 0.986 & 0.995 \\
Lee \etal (BTS)~\cite{lee2019big} & $0 - 80\ m$ & \textbf{0.064} & \textbf{0.254} & \textbf{2.815} & \textbf{0.100} & \textbf{0.950} & \textbf{0.993} & \textbf{0.999} \\

\bottomrule
\end{tabular}%
} 


\end{table*}

Besides evaluating the FCRN architecture using different versions of the KITTI datasets, we conducted some ablation studies to identify the best training combination. More specifically, we used different loss functions and studied the influence of using all pixels, which includes the sky and reflecting surfaces, or only valid pixels, which have corresponding depth information. In the conducted experiments, the best models were consistently obtained when using the BerHu loss, what reaffirms the conclusions presented by~\cite{Laina2016}, and when using only valid information.

In Table~\ref{tab:sparsity_analysis_eigen_kitti_depth_split_results}, we present the results obtained from our method when trained on the proposed datasets and evaluated on the \textit{Eigen Split based on KITTI Depth}. The models trained on the \textit{KITTI Continuous} presented a better performance than those trained on \textit{KITTI Depth}, which confirms our initial hypothesis which stated that ones trained on continuous maps had the potential of outperforming those trained on semi-dense dataset. Since all models have the same network architecture, these gains in performance demonstrate how the densification process produced by the Hilbert Maps framework can improve the quality of predictions when compared to baseline versions (without HM).
The same technique can be applied to other network architectures, to produce similar performance gains that can push current state-of-the-art models even further. Moreover, we show how misleading results obtained from semi-dense data can be when applied to dense benchmarks.

\begin{figure*}[htp]
\centering
\vspace{2mm}
\includegraphics[width=\textwidth]{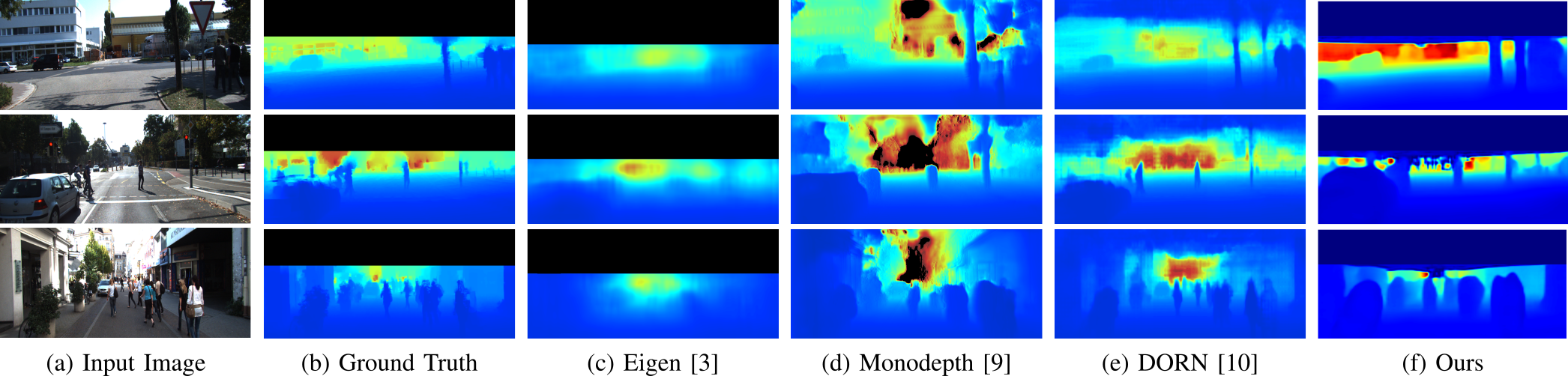}
\caption{Qualitative comparison between our estimates and other state-of-art works.}
\label{fig:kitti_compare_soa}
\end{figure*}

The quantitative and qualitative comparisons between our results and the current state-of-the-art are presented in Table~\ref{tab:sparsity_analysis_eigen_kitti_depth_split_results} and Figure~\ref{fig:kitti_compare_soa}, respectively. Even that the chose architecture predicts meaningful depth estimates when trained on \textit{KITTI Depth}, numerically, its performance does not please. We believe that its inferior performance is related to the suitability of the architecture for processing sparse information since it was originally projected for \textit{NYU Depth v2} dataset (dense) and doesn't fail on the proposed in our \textit{KITTI Continuous} dataset. Like DORN~\cite{Fu2018}, our method also detects well obstacles present in the scenes, with the noticeable difference that ours provide a certain margin of safety around the obstacles (artifacts). They are caused due to the reconstructive properties of the Hilbert Maps framework, as describled in section~\ref{sec:occupancy_maps}.

\begin{figure*}[htp]
\centering
\includegraphics[width=0.95\textwidth]{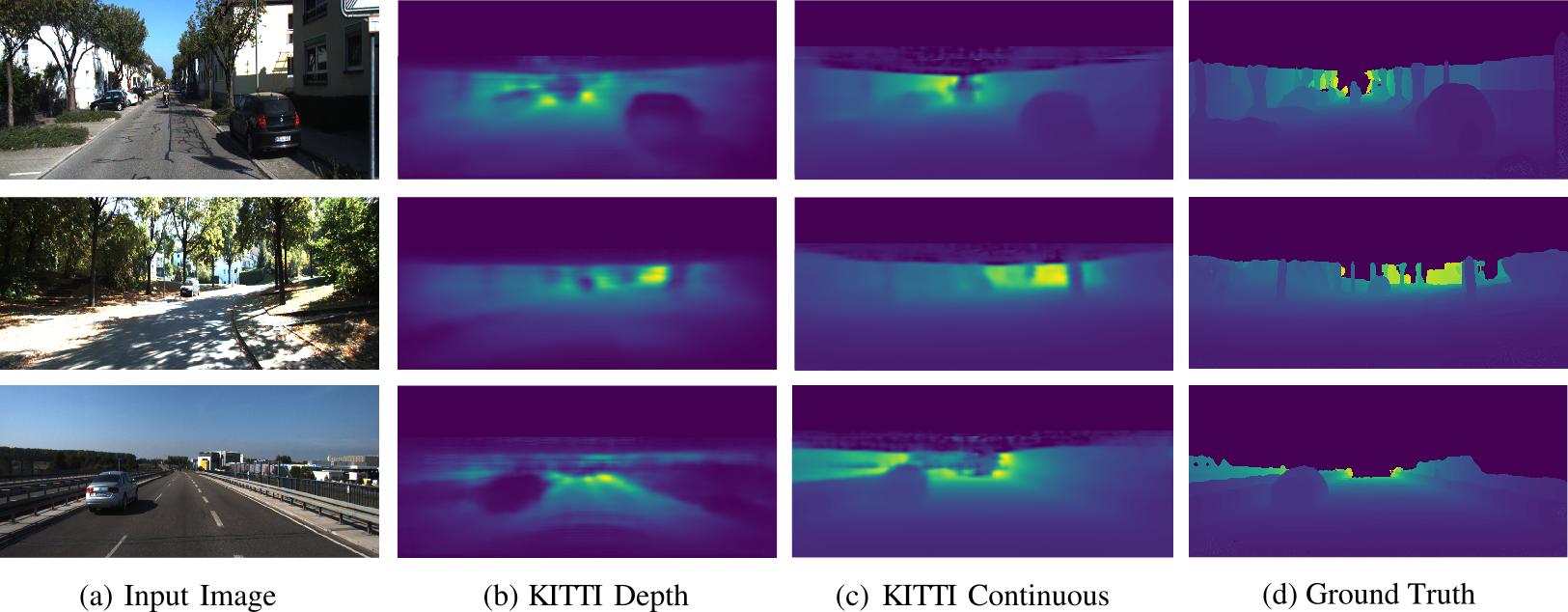}
\caption{Qualitative comparison between depth predictions when trained on the proposed datasets. (a) Input RGB Image. (b) KITTI Depth (Semi-Dense, 11 Scans). (c) KITTI Continuous (Dense, Continuous Occupancy Maps). (d) Ground Truth (Continuous).}
\label{fig:kitti_compare}
\end{figure*}

Figure~\ref{fig:kitti_compare} illustrates the qualitative comparison between the predictions when training on the proposed datasets. As can be noted, the continuous depth images boosted up the quality of distance estimations. In other words, they make the predicted images much less blurred, i.e., they have a better definition of the edges of the objects, also having more accurate measurements according to the ground truth maps. The main cause of the predictions of sparse datasets to be blurred is the use of 2D convolutional filters in widely sparse regions and the occasionality of depth information since the distance value in a given pixel is intermittent and this depends on where the laser points will be reprojected.

%% file: 05conclusion.tex
\section{Conclusion}
\label{sec:conclusion}

In this paper, we present a monocular depth estimation algorithm that uses Hilbert Maps, whose occupancy models were used as a preprocessing step to the \textit{Single Image Depth Estimation} problem. These models generate continuous depth maps for training a deep residual network, differing from typical supervised approaches that use sparse ones. To the best of our knowledge, this work is the first to leverage a 3D reconstruction tool as in-painting to improve depth estimates. Furthermore, the proposed idea is not restricted to these techniques alone and does not require any other type of sensors or extra information, only RGB images as input and continuous depth maps as supervision, which significantly improved the quality of network predictions over typical sparse maps. Moreover, the proposed methodology presented superior performance (First and second rows in Table~\ref{tab:sparsity_analysis_eigen_kitti_depth_split_results}) even when using 67.6\% fewer examples than those trained in the \textit{KITTI Depth} dataset, as a consequence of increasing the valid information present in the ground truth maps from 45.98\% to 64\%. The main limitations of the proposed preprocessing method are the computational cost required to compute each continuous depth map used for training, and that it is bounded by the reconstructed capabilities of the Hilbert Maps framework.

Future work will focus on optimizing the method itself, mainly tackling the aforementioned limitations, and honing the network topology by incorporating new architectural developments that are more suited to the SIDE task. Further investigation of other depth-only variants of in-painting techniques~\cite{doria2012filling, Gong2013, Chen2014inpainting, Lu2015, Uhrig2017} for the proposed preprocessing step will also be conducted, for an evaluation of how Hilbert Maps comparably performs relative to them. Further analysis of how color distortion may influence the depth estimates can also be suggested.

%% file: 06acknowledgment.tex
\section*{Acknowledgment}

This research was supported by funding from the Brazilian National Council for Scientific and Technological Development (CNPq), under grant 130463/2017-5 and 465755/2014-3, financed in part by the Coordination of Improvement of Higher Education Personnel - Brazil - CAPES (Finance Code 001 and 88887.136349/2017-00), the São Paulo Research Foundation (FAPESP) grant 2014/50851-0, and the Faculty of Engineering \& Information Technologies, University of Sydney (USYD). We also gratefully acknowledge the support of NVIDIA Corporation with the donation of the Titan X GPUs used on this research.

%% file: 07references.tex
\bibliographystyle{IEEEtran}
